\documentclass[sigconf]{acmart}
\AtBeginDocument{%
  }

\setcopyright{acmlicensed}
\copyrightyear{2025}
\acmYear{2025}
\acmDOI{}

\acmConference[SC'25]{The International Conference for High Performance Computing, Networking, Storage and Analysis}{November 16--21, 2025}{St. Louis, MO, USA}

\acmISBN{}
\usepackage{algorithm}
\usepackage{algorithmic}
\usepackage{subcaption}
\usepackage{multirow} 
\usepackage{booktabs}
\usepackage{graphicx}

\begin{document}

\title{Generative Latent Diffusion for Efficient Spatiotemporal Data Reduction}

\author{Xiao Li}
\affiliation{%
  \institution{University of Florida}
  \city{Gainesville}
  \state{FL}
  \country{USA}
}

\author{Liangji Zhu}
\affiliation{%
  \institution{University of Florida}
  \city{Gainesville}
  \state{FL}
  \country{USA}
}

\author{Anand Rangarajan}
\affiliation{%
  \institution{University of Florida}
  \city{Gainesville}
  \state{FL}
  \country{USA}
}

\author{Sanjay Ranka}
\affiliation{%
  \institution{University of Florida}
  \city{Gainesville}
  \state{FL}
  \country{USA}
}

\renewcommand{\shortauthors}{Xiao Li, Liangji Zhu, Anand Rangarajan, Sanjay Ranka}

\begin{abstract}

Generative models have demonstrated strong performance in conditional settings and can be viewed as a form of data compression, where the condition serves as a compact representation. However, their limited controllability and reconstruction accuracy restrict their practical application to data compression. In this work, we propose an efficient latent diffusion framework that bridges this gap by combining a variational autoencoder with a conditional diffusion model. Our method compresses only a small number of keyframes into latent space and uses them as conditioning inputs to reconstruct the remaining frames via generative interpolation, eliminating the need to store latent representations for every frame. This approach enables accurate spatiotemporal reconstruction while significantly reducing storage costs. Experimental results across multiple datasets show that our method achieves up to 10× higher compression ratios than rule-based state-of-the-art compressors such as SZ3, and up to 63\% better performance than leading learning-based methods under the same reconstruction error.

\end{abstract}


\keywords{Latent Diffusion Models, Data Reduction, Spatiotemporal Scientific Application}

\maketitle

\section{Introduction}
Current scientific simulations generate massive, high-resolution spatiotemporal datasets that stress existing storage pipelines \cite{MGARD_1,xiao_climate2,xiao_bigdata}. Traditional compression techniques\cite{sz3,zfp}, while effective in general-purpose settings, often fall short in preserving the structural and statistical nuances required for scientific analysis. A key requirement in this domain is the ability to enforce error-bound guarantees, particularly with respect to primary data (PD) and quantities of interest (QoI), ensuring that downstream scientific analysis remains valid after compression. 

Prior work \cite{xiao_bigdata} has made substantial progress in this direction, introducing foundation models for scientific data compression based on attention mechanisms over tensor hyper-blocks. These models are capable of learning latent space embeddings of tensor block representations that satisfy strict error constraints on PD and QoI, and have demonstrated strong performance across a range of scientific applications. However, these efforts have been primarily transform-based, relying on autoencoders, quantization, and entropy coding applied independently to spatial or spatiotemporal tensor blocks.

In this work, we take a significant step forward by integrating \textbf{transform-based} compression with \textbf{temporal interpolation} across blocks, leveraging generative modeling to enhance compression. Our approach combines a variational autoencoder (VAE) with a hyperprior module to compress selected keyframes of data blocks, producing compact latent space embeddings. These embeddings are then used not only to reconstruct the keyframes themselves but also to guide a conditional diffusion (CD) model trained to interpolate missing blocks across time. Specifically, the CD model starts from a noisy input (except for the keyframes themselves) and progressively performs denoising to generate plausible intermediate frames, with the latent embeddings of neighboring keyframes available as side information.

This hybrid approach enables a novel form of generative interpolation in latent space, allowing temporal gaps to be filled in a statistically coherent, data-driven manner. By using generative AI not for synthesis but for efficient interpolation, our method repurposes diffusion models as learned, nonlinear interpolators (akin to higher-dimensional splines) that serve the broader goal of data reduction with fidelity guarantees. Crucially, to this end, we include a post-processing module that ensures error-bound guarantees on the final reconstruction, supporting scientific use cases that demand quantifiable accuracy of the PD. We believe this work opens new possibilities for applying generative modeling to compression, thereby aiding how scientific data is stored, processed, and shared.

We evaluate the proposed method across three scientific domains: E3SM \cite{e3sm}, S3D \cite{s3d}, and JHTDB \cite{jhtdb}, and compare its performance against both rule-based and learning-based state-of-the-art compression approaches. Experimental results demonstrate that our method achieves 4--10$\times$ higher compression ratios at the same reconstruction error when compared to rule-based methods, and delivers improvements of 20\% to 63\% over leading learning-based compressors.

The key contributions of this work are summarized as follows:
\begin{itemize}
    \item We propose a novel generative compression framework based on latent diffusion, which stores only keyframes and reconstructs intermediate frames through learned spatiotemporal interpolation.
    \item To improve the efficiency and practicality of the method, we fine-tune the diffusion model to operate with significantly fewer denoising steps, enabling its use in real-time or large-scale scientific data compression scenarios.
    \item Our approach includes a post-processing module that guarantees reconstruction error bounds, making it suitable for scientific workflows requiring quantifiable accuracy on primary data.
    \item Extensive evaluations on multiple scientific domains show that our method consistently outperforms existing baselines, achieving up to 63\% improvement over state-of-the-art learning-based compressors.
\end{itemize}

\section{Related Work}
Error-bounded lossy compression is widely recognized as one of the most effective approaches for scientific data reduction, primarily due to its ability to guarantee reliability through strict error control. This ensures that the compression process preserves data fidelity within a user-defined tolerance, which is crucial in scientific computing. Broadly, lossy compression methods fall into two categories: traditional method and learning-based method.

\paragraph*{Rule-Based Methods}

Rule-based compression methods rely on mathematical models to reduce data size without training on large datasets. ZFP \cite{fox2020stability} compresses data by dividing it into non-overlapping 4D blocks and applying a near-orthogonal transform to decorrelate values efficiently. TTHRESH \cite{ballester2019tthresh} uses higher-order singular value decomposition (HOSVD) to reduce dimensionality, retaining only the most significant components of the dataset. MGARD \cite{MGARD_2, gong2023mgard} applies a multigrid-based approach to transform floating-point data into a hierarchy of coefficients, enabling progressive resolution and rigorous error control. Several traditional techniques estimate data values to improve compressibility. For example, Differential Pulse Code Modulation (DPCM) \cite{mun2012dpcm} encodes the difference between successive values, while MPEG standards \cite{boyce2021mpeg} employ interpolative techniques to predict samples from their neighbors. SZ \cite{SZ_3}, a widely adopted method in scientific computing, enhances prediction quality by leveraging adjacent data points and supports multiple variations tailored to different use cases. FAZ \cite{liu2023faz} introduces a modular framework that combines prediction schemes and wavelet transforms for robust and adaptable compression. These methods are typically very fast but tend to achieve relatively lower compression ratios compared to learning-based approaches.

\paragraph*{Learning-Based Methods}

Machine learning techniques have recently demonstrated strong potential in lossy compression, especially in preserving high-fidelity reconstructions. Variational Autoencoders (VAEs) \cite{kingma2013auto}, along with various enhancements \cite{minnen2018joint, kingma2021variational}, have significantly advanced the field. For instance, \cite{minnen2018joint} combines autoregressive models with hierarchical priors, enabling context-aware prediction of latent variables guided by a hyperprior-informed autoencoder. \cite{xiaofm} extend the VAE framework by incorporating super-resolution modules and alternating 2D and 3D convolutions to effectively capture spatiotemporal correlations, achieving both high compression ratios and high-fidelity reconstructions. \cite{xiaoli_atten} propose a block-wise compression strategy that leverages self-attention mechanisms to better capture local and inter-block dependencies within the data.

Diffusion models have recently achieved remarkable progress in image generation tasks \cite{diffusion_survey,diffusion_survey2}. Conditional diffusion models \cite{voleti2022MCVD,zhan2024conditional}, in particular, have been explored for their ability to guide the generation process using auxiliary information, which can be treated as compressed representations. However, unlike general image generation, where realism is the primary objective, data compression demands that the reconstructed output closely matches the original data. This strict fidelity requirement renders naive random generation unsuitable for compression tasks. To address this challenge, recent works \cite{relic2024lossy,cdc} propose a conditional diffusion framework that encodes images into a latent space and reconstructs them using the latent embeddings as conditioning inputs. Building on this idea, GCD \cite{gcd} extends the approach from 2D images to 3D data by employing a block-based strategy.

However, all these methods require storing the latent representation for every frame or evey block, which leads to inefficiencies in storage. To address this, we propose a more efficient approach that retains only the latent space of keyframes. These key latents serve as conditioning inputs to generate the latent representations of intermediate frames. A latent diffusion model is employed to perform this generation process. Our method demonstrates significantly better results in terms of both compression ratio and reconstruction fidelity compared to prior approaches. In the following section, we introduce each component of our proposed method in detail.

\section{Method}
\begin{figure*}[htbp]
    \centering
    \includegraphics[width=\textwidth]{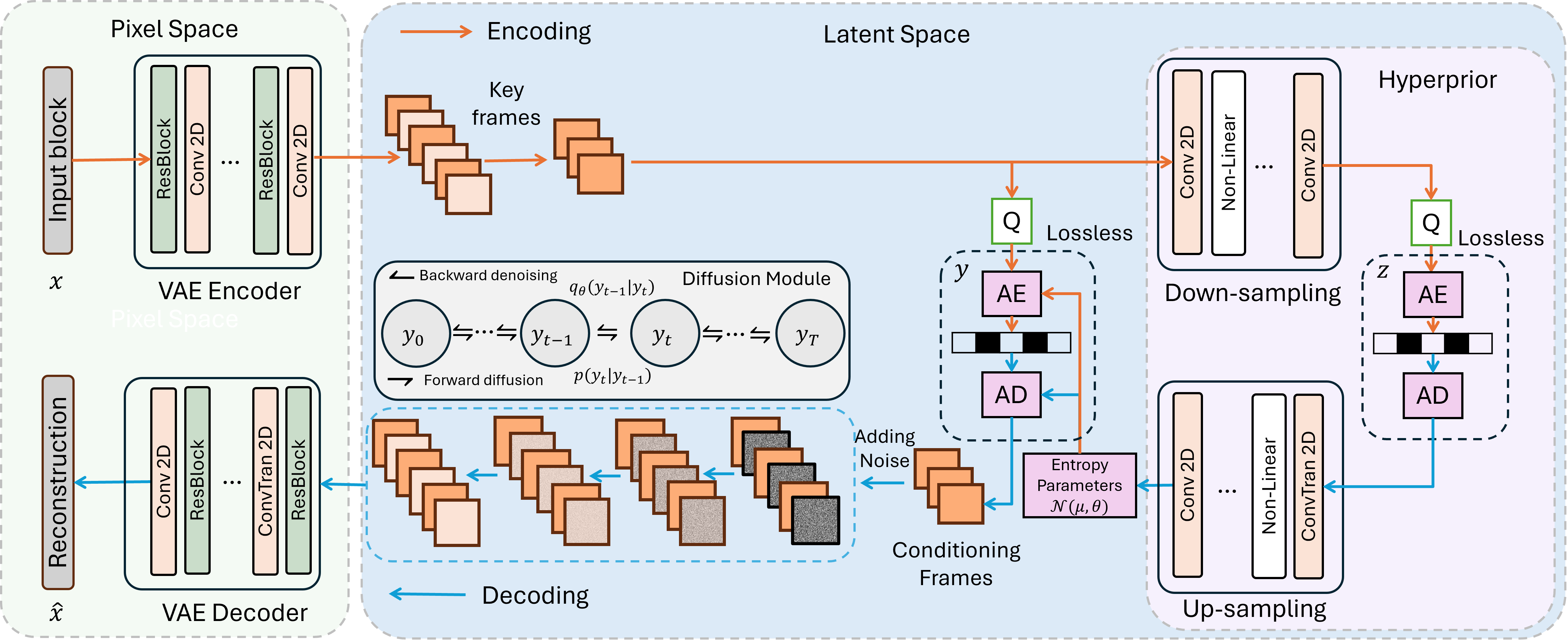}  
    \caption{Overview of the proposed architecture. Given an input spatiotemporal data block $x$, each frame is encoded into a latent space using a VAE encoder. Only the latent representations of selected keyframes are retained, quantized, and losslessly compressed using a hyperprior module. These keyframe latents serve as conditioning inputs for a latent diffusion model, which generates the latent representations of the non-keyframes through interpolation. Finally, both the keyframe and generated non-keyframe latents are decoded by the VAE decoder to reconstruct the full data block $\hat{x}$.}

    \label{fig:overview}
\end{figure*}

Conventional methods store latent representations for every frame, limiting compression efficiency. In contrast, we improve the compression ratio by employing a generative approach with a latent diffusion model. Instead of storing latents for all frames, we retain only the keyframe latents, which serve as conditions for the diffusion model to generate latent representations for other frames.

Our framework consists of three key components: a Variational Autoencoder (VAE), a hyperprior module, and a latent diffusion module, as illustrated in Figure \ref{fig:overview}. The encoder first compresses keyframes into a compact latent space, while the hyperprior module models the latent distribution for conditional entropy coding. The latent diffusion module then propagates information by generating latent representations for non-key frames. Finally, the decoder reconstructs the full temporal sequence from these latents. We introduce each component below.

\subsection{VAE with Hyperprior}
When using transform-based compression, the encoder \(\mathcal{E}_x\) computes the latent space \(\mathcal{E}_x(\boldsymbol{x})\) from the input data \(\boldsymbol{x}\). To enhance compression efficiency, the latent space is quantized by rounding, yielding \(\boldsymbol{y} = \text{Round}(\mathcal{E}_x(\boldsymbol{x}))\), and then further compressed using lossless entropy coding. The decoder \(\mathcal{D}_x\) reconstructs the data as \(\hat{\boldsymbol{x}} = \mathcal{D}_x(\boldsymbol{y})\). Compression performance is evaluated by distortion \(D\), which measures the difference between \(\boldsymbol{x}\) and \(\hat{\boldsymbol{x}}\), and bit-rate \(R\), which represents the number of bits required to encode \(\boldsymbol{y}\).

To optimize latent space compression, \cite{minnen2018joint} models each element \(\boldsymbol{y}_i\) as a Gaussian distribution \(\mathcal{N}(\boldsymbol{\mu}_i, \boldsymbol{\sigma}_i^2)\), with mean \(\boldsymbol{\mu}_i\) and variance \(\boldsymbol{\sigma}_i^2\). A hyperprior autoencoder (AE) captures the latent space distribution using a hyperencoder \(\mathcal{E}_h\) and hyperdecoder \(\mathcal{D}_h\). The hyperlatent space \(\boldsymbol{z} = \mathcal{E}_h(\boldsymbol{y})\) is quantized to \({\boldsymbol{z}}\), and \(\mathcal{D}_h({\boldsymbol{z}})\) estimates \((\boldsymbol{\mu}, \boldsymbol{\sigma})\) for \(\boldsymbol{y}\). Each quantized element \(\boldsymbol{y}_i\) is then modeled as:
\begin{equation}
    \boldsymbol{y}_i \sim \mathcal{N}(\boldsymbol{\mu}_i, \boldsymbol{\sigma}_i^2) * \mathcal{U}\left( -0.5, 0.5 \right),
\end{equation}
where \(\mathcal{U}\left( -0.5, 0.5 \right)\) represents the uniform distribution introduced by the quantization process, with \( * \) denoting convolution. The conditional probability \(p({\boldsymbol{y}} \mid \boldsymbol{\mu}, \boldsymbol{\sigma})\) is then given by:
\begin{equation}
    p(\boldsymbol{y} \mid \boldsymbol{\mu}, \boldsymbol{\sigma}) = \prod_i \left( \mathcal{N}(\boldsymbol{\mu}_i, \boldsymbol{\sigma}_i^2) * \mathcal{U}\left( -0.5, 0.5 \right) \right)(\boldsymbol{y}_i).
\end{equation}
Since there is no prior belief about the hyperlatent space, \({\boldsymbol{z}}\) is modeled with the non-parametric, fully factorized density model \(p({\boldsymbol{z}})\) \cite{vae_z}. Entropy models, such as arithmetic coding \cite{arithmetic}, can losslessly compress \({\boldsymbol{y}}\) given the probability model. Therefore, the bit-rate for \({\boldsymbol{y}}\) can be expressed as \( R_{{y}} = \mathbb{E}_{{\boldsymbol{y}}} \left( -\log_2(p({\boldsymbol{y}} \mid \boldsymbol{\mu}, \boldsymbol{\sigma})) \right) \). The bit-rate for \({\boldsymbol{z}}\) can be expressed as \( R_{{z}} = \mathbb{E}_{{\boldsymbol{z}}} \left( -\log_2(p({\boldsymbol{z}} )) \right) \). The total bit-rate can be calculated as $R = R_{{y}} + R_{{z}} $.

\subsection{Latent Diffusion Models}

\textit{Latent diffusion models} \cite{rombach2022high,blattmann2023align} are probabilistic frameworks designed to learn a latent data distribution \( p(y) \) by progressively denoising a Gaussian noise variable. This corresponds to modeling the reverse process of a fixed Markov chain with length \( T \). The forward process introduces noise step by step, following a Gaussian transition:
\begin{equation}
q(\boldsymbol{y}_t | \boldsymbol{y}_{t-1}) = \mathcal{N}(\boldsymbol{y}_t | \sqrt{1 - \beta_t} \boldsymbol{y}_{t-1}, \beta_t \boldsymbol{I}),
\end{equation}
where \( \beta_t \) is a noise schedule controlling the variance at each step. Additionally, the forward process can be described as transitioning from the initial data \( x_0 \) to the noisy data \( x_t \) using:
\begin{equation}
q(\boldsymbol{y}_t | \boldsymbol{y}_0) = \mathcal{N}(\boldsymbol{y}_t | \sqrt{\bar{\alpha}_t} \boldsymbol{y}_0, (1-\bar{\alpha}_t) \boldsymbol{I}),
\end{equation}
where \( \bar{\alpha}_t = \prod_{i=1}^t (1 - \beta_i) \) is the cumulative product of the noise schedule, determining the amount of noise added by time step \( t \). The reverse process, parameterized by a neural network \( \epsilon_{\theta} \), aims to recover the clean data by estimating the added noise. For image synthesis, state-of-the-art models \cite{kwon2022diffusion,blattmann2023stable} optimize a reweighted variational lower bound on \( p(\boldsymbol{y}) \), aligning with denoising score matching \cite{ho2022video}. These models can be interpreted as a sequence of denoising autoencoders \( \epsilon_{\theta}(\boldsymbol{y}_t, t) \) trained to predict a noise-free version of \( \boldsymbol{y}_t \). The training objective simplifies to:

\begin{equation}
L_{\text{DM}} = \mathbb{E}_{\boldsymbol{y}, \epsilon \sim \mathcal{N}(0,1), t} \left[ \|\epsilon - \epsilon_{\theta}(\boldsymbol{y}_t, t)\|_2^2 \right],
\end{equation}

where \( t \) is uniformly sampled from \(\{1, \dots, T\}\).
By iteratively applying the learned denoising process, the model reconstructs high-fidelity data from random noise, making diffusion models highly effective for generative tasks.

\paragraph{Denoising UNet}

Our denoising UNet is adapted from the architecture proposed in \cite{ho2022video}, originally developed for video generation in the pixel space. It utilizes factorized space-time attention to efficiently capture spatiotemporal correlations. This attention mechanism has been shown to be both effective and computationally efficient in video transformers \cite{arnab2021vivit,bertasius2021space}. Specifically, given an input tensor \( X \in \mathbb{R}^{N \times H \times W \times C} \), where \( N \) is the number of frames (temporal dimension), \( H \) and \( W \) are the spatial height and width of each frame, and \( C \) is the number of feature channels, temporal attention is applied by reshaping the input to \( (H \cdot W) \times N \times C \) and computing self-attention along the temporal dimension. Spatial attention is applied by reshaping to \( N \times (H \cdot W) \times C \) and using the same attention formula within each frame. We extend this model \cite{ho2022video} to operate in the latent diffusion setting by changing the input and output channels from 3 to 64, corresponding to the dimensionality of the latent space produced by our VAE.

\subsection{Keyframe Conditioning}
\label{sec:condition}
Let \( \boldsymbol{y}^{\mathcal{N}}_0 \in \mathbb{R}^{N,C,W,H} \) be the quantized latent representation of \( N \) frames, where \( C \), \( W \), and \( H \) denote the number of channels, width, and height of the latent features, respectively. We normalize \( \boldsymbol{y}^{\mathcal{N}}_0 \) to the range \([-1, 1]\) by applying min-max normalization. We observe that learning degrades when the latent dynamic range is misaligned with the noise scale. For simplicity, we continue to denote the normalized latent space as \( \boldsymbol{y}^{\mathcal{N}}_0 \). We further partition \( \boldsymbol{y}^{\mathcal{N}}_0 \) into two subsets along the temporal dimension: the frames to be generated \( \boldsymbol{y}^{\mathcal{G}}_0 \in \mathbb{R}^{N-K,C,W,H} \) and the conditioning frames \( \boldsymbol{y}^{\mathcal{C}}_0 \in \mathbb{R}^{K,C,W,H} \), where \( K \) is the number of conditioning frames, and \( \mathcal{G} \) and \( \mathcal{C} \) are index sets such that \( \mathcal{G} \cap \mathcal{C} = \emptyset \) and \( \mathcal{G} \cup \mathcal{C} = \mathcal{N} = \{1, 2, \dots, N\} \). We express \( \boldsymbol{y}^{\mathcal{N}}_0 \) as \( \boldsymbol{y}^{\mathcal{G}}_0 \oplus \boldsymbol{y}^{\mathcal{C}}_0 \) with the following definition for \( \oplus \):

\[
(a^{\mathcal{G}} \oplus b^{\mathcal{C}})^i :=
\begin{cases}  
a^i, & \text{if } i \in \mathcal{G} \\  
b^i, & \text{if } i \in \mathcal{C}  
\end{cases}
\]
Given the conditional information \( \boldsymbol{y}^{\mathcal{C}} \), conditional diffusion models usually optimize:

\begin{equation}
    L_{\text{CDM}} = \mathbb{E}_{\boldsymbol{y}^{\mathcal{G}}, \boldsymbol{y}^{\mathcal{C}}, \epsilon \sim \mathcal{N}(0,1), t} \left[ \|\epsilon - \epsilon_{\theta}(\boldsymbol{y}^{\mathcal{G}}_t, \boldsymbol{y}^{\mathcal{C}}_0, t)\|_2^2 \right].
\end{equation}

Some works condition the model using a pseudo-prompt and apply cross-attention mechanisms to the U-Net architecture \cite{blattmann2023stable}, or alternatively, provide the conditioning as a separate input and use an additional layer for fusion \cite{van2016conditional}. On the other hand, we give the entire sequence to the denoising network \( \epsilon_{\theta} \), only adding noise to the frame that is to be generated. The forward diffusion process is given by $\boldsymbol{y}^{\mathcal{G}}_t \sim \mathcal{N}(\boldsymbol{y}^{\mathcal{G}}_t | \sqrt{\bar{\alpha}_t} \boldsymbol{y}^{\mathcal{G}}_0, (1-\bar{\alpha}_t) I).$

Further, we aggregate the noisy data \( \boldsymbol{y}^{\mathcal{G}}_t \) with the conditioning frame \( \boldsymbol{y}^{\mathcal{C}}_0 \) to form the input to the network, \( \boldsymbol{y}^{\mathcal{N}}_t = \boldsymbol{y}^{\mathcal{G}}_t \oplus \boldsymbol{y}^{\mathcal{C}}_0 \). The network outputs an estimated noise with the same dimensions as the original data. We then use the corresponding frame output to estimate the noise of the input. However, we are only concerned with the noise of the frame to be generated, denoted as \( \epsilon_{\theta}(\boldsymbol{y}^{\mathcal{N}}_t, t)^{\mathcal{G}} \). The loss is computed only with respect to \( \boldsymbol{y}_t^{\mathcal{G}} \):

\begin{equation}
    L_{\text{CDM}} = \mathbb{E}_{\boldsymbol{y}^{\mathcal{N}}, \epsilon \sim \mathcal{N}(0,1), t} \left[ \| \epsilon - \epsilon_{\theta}(\boldsymbol{y}^{\mathcal{N}}_t, t)^{\mathcal{G}} \|_2^2 \right],
\end{equation}
where \( \epsilon \) has the same dimensions as \( \boldsymbol{y}_t^{\mathcal{G}} \).



\subsection{Model Training}

The training process comprises two stages. In the first stage, a variational autoencoder (VAE) with a hyperprior module is trained to compress 2D images into a compact latent representation. Once trained, the VAE encoder is frozen. In the second stage, a latent diffusion model is trained using the output of the frozen encoder as input.

\paragraph{VAE with hyperprior Training}

The training objective comprises two terms: distortion \( D \) and bit-rate \( R \). Distortion is measured using the Mean Squared Error (MSE) between the original data \( \boldsymbol{x} \) and its reconstruction \( \hat{\boldsymbol{x}} \). The bit-rate term accounts for both the latent representation \( {\boldsymbol{y}} \) and the hyper-latent representation \( {\boldsymbol{z}} \), where a Lagrange multiplier \( \lambda \) balances the trade-off between compression efficiency and reconstruction fidelity. The overall loss function is given by:  

\begin{equation}
    L = \text{MSE}(\boldsymbol{x}, \hat{\boldsymbol{x}}) + \lambda \left( \mathbb{E}_{{\boldsymbol{y}}} \left[ -\log_2 p({\boldsymbol{y}} \mid \boldsymbol{\mu}, \boldsymbol{\sigma}) \right] + \mathbb{E}_{{\boldsymbol{z}}} \left[ -\log_2 p({\boldsymbol{z}}) \right] \right).
\end{equation}

Quantization cannot be directly applied during training due to its non-differentiable nature. To address this, uniform noise \( \mathcal{U}(-0.5, 0.5) \) is added to the latent variables \( \boldsymbol{y} \) and \( \boldsymbol{z} \) during the forward pass to approximate quantization while preserving differentiability. At inference time, actual quantization is performed by rounding to the nearest integer. This approach enables end-to-end training and accurate entropy modeling within variational compression frameworks.

\paragraph{Diffusion Model Training}

Since the dimensionality of \( {\boldsymbol{y}} \) is significantly smaller than that of \( {\boldsymbol{z}} \), the memory overhead of storing \( {\boldsymbol{z}} \) is often negligible. As a result, we primarily focus on the generation and efficient representation of \( {\boldsymbol{y}} \). To train the diffusion model, we randomly sample $N$ consecutive frames from the original spatio-temporal data block. Each of N frames are then processed separately by the pre-trained VAE encoder $\mathcal{E}_x$ followed by rounding quantization followed by normalization, which projects them into the latent space. As described in Section~\ref{sec:condition}, we construct the input for the diffusion model by introducing noise to the frames designated for generation while keeping the conditioning frames intact. The model learns to iteratively denoise the generated frames while leveraging temporal dependencies captured in the conditioning frames. The training process is summarized in Algorithm \ref{alg:train}.

\begin{algorithm}[h]
    \caption{Training Latent Diffusion Model}
    \begin{algorithmic}[1]
        \STATE Given frozen pre-trained encoder \( \mathcal{E}_x \) and denoising UNet \( \epsilon_{\theta} \)
        \REPEAT
            \STATE Sample \( N \) consecutive frames from the original data block, denoted as \(\boldsymbol{x}^{\mathcal{N}} \)
            
            \STATE \( \boldsymbol{y}^{i} \gets Round(\mathcal{E}_x(\boldsymbol{x}^{i})) \) \hfill \textit{(Project each frame to latent space and apply quantization)}
            \STATE \( \boldsymbol{y}^{\mathcal{N}}_0 \gets \text{Min-Max Normalize}(\boldsymbol{y}^{\mathcal{N}}) \)
            \STATE \( (\boldsymbol{y}^{\mathcal{C}}_0, \boldsymbol{y}^{\mathcal{G}}_0) \gets \text{Partition}(\boldsymbol{y}^{\mathcal{N}}_0) \)
            \STATE Sample \( t \sim \text{Uniform}(1, T) \)
            \STATE \( \epsilon \sim \mathcal{N}(\boldsymbol{0}, \boldsymbol{I}) \)
            \STATE \( \boldsymbol{y}^{\mathcal{G}}_{t} \gets \sqrt{\bar{\alpha}_t} \boldsymbol{y}^{\mathcal{G}}_0 + \sqrt{1 - \bar{\alpha}_t} \epsilon \)
            \STATE \( \boldsymbol{y}_t^{\mathcal{N}} \gets \boldsymbol{y}^\mathcal{G}_{t} \oplus \boldsymbol{y}_{0}^{\mathcal{C}} \) 
            \STATE \( \hat{\epsilon} \gets \epsilon_{\theta}(\boldsymbol{y}_t^{\mathcal{N}}, t) \) \hfill \textit{(Estimate noise)}
            \STATE \( L_{\text{CDM}} \gets \|\epsilon - \hat{\epsilon}^{\mathcal{G}}\|_2^2 \) \hfill \textit{(Compute loss)}
            \STATE Update model parameters $\theta$ using \( \nabla L_{\text{CDM}} \)
        \UNTIL{convergence}
        \STATE Save trained model \( \epsilon_{\theta} \)
    \end{algorithmic}
    \label{alg:train}
\end{algorithm}

\subsection{Error Bound Guarantee}

In scientific data compression, ensuring a guaranteed error bound is essential to maintain the accuracy of the reconstructed data, especially when dealing with large and complex datasets. Following the work of \cite{xiaofm, jaemoon2,jaemoon3,xiao_climate2}, we employ a PCA-based method to achieve this guarantee. After compressing and decompressing the data, we apply PCA to the residuals to extract the principal components, yielding the basis matrix \(\boldsymbol{U}\). The residual is then projected onto \(\boldsymbol{U}\), and the coefficients \(\boldsymbol{c}\) are computed:

\begin{equation}
    \boldsymbol{c} = \boldsymbol{U}^{T}(\boldsymbol{x} - \boldsymbol{x}^R).
\end{equation}
To satisfy the error bound, we select the top \(M\) coefficients based on their contribution to the error such that the \(\ell_2\)-norm of the residual falls below a specified threshold \(\tau\), i.e., \(\left\|\boldsymbol{x} - \boldsymbol{x}^G\right\|_{2} \leq \tau\). These coefficients are quantized and entropy-coded to minimize storage cost. The corrected reconstruction is then given by:

\begin{equation}
\boldsymbol{x}^{G} = \boldsymbol{x}^{R} + \boldsymbol{U}_{s}\boldsymbol{c}_{q},
\end{equation}

where \(\boldsymbol{c}_{q}\) represents the quantized set of selected coefficients, and \(\boldsymbol{U}_{s}\) denotes the corresponding set of basis vectors.

\section{Experimental Results}
In this section, we begin by introducing the evaluation metrics and datasets used in our study. We then explore the conditioning strategies that yield the best performance for data compression, followed by a discussion of the hyperparameters used in our experiments. Next, we compare our method with state-of-the-art approaches in terms of compression ratio and error curves. Finally, we evaluate computational efficiency of our method and compare with other diffusion-based methods.

\begin{figure*}[t]
  \centering
  \begin{subfigure}[b]{\textwidth}
    \centering
    \includegraphics[width=\textwidth]{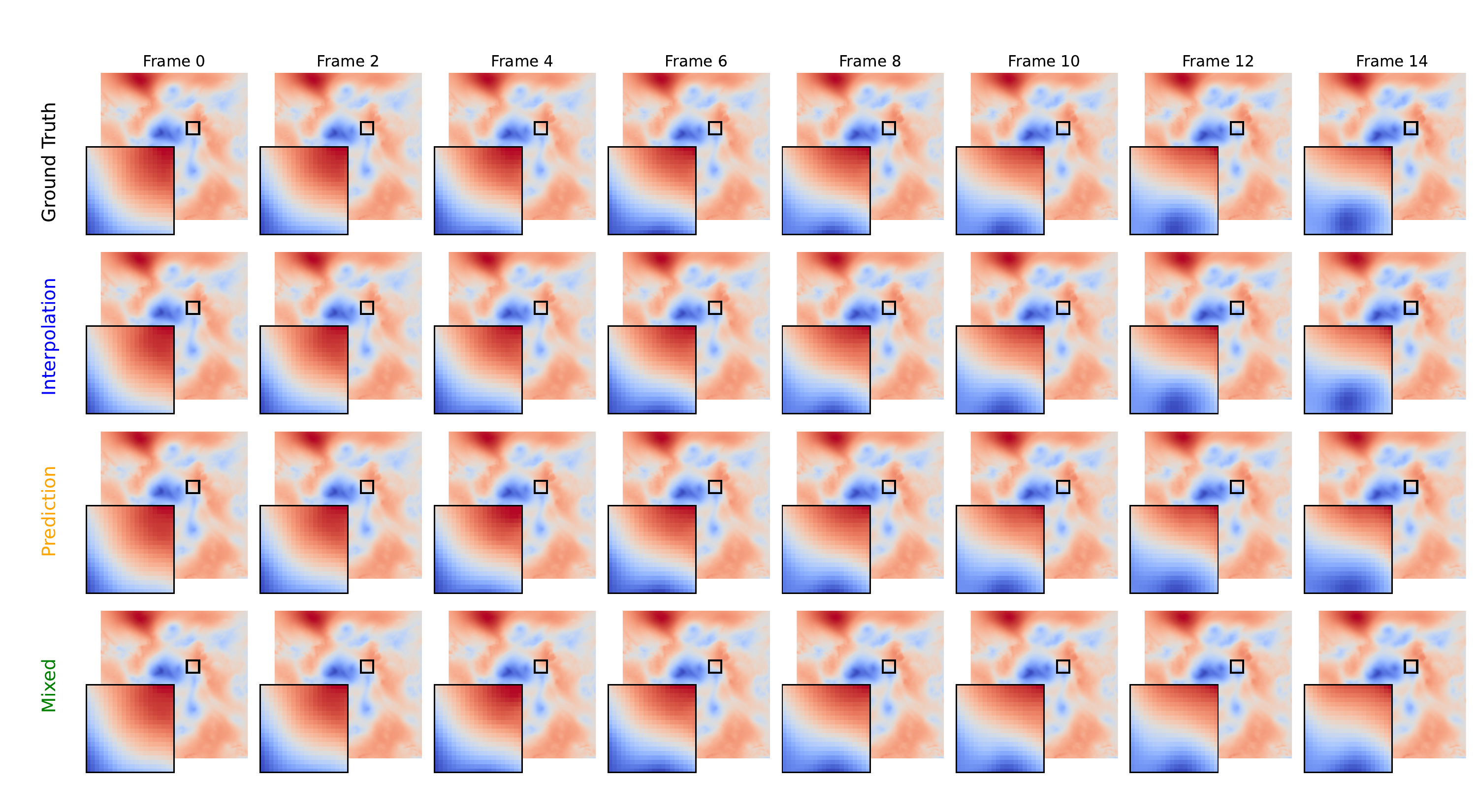}
    \label{fig:compare_visual}
  \end{subfigure}
  \vspace{-0.8cm}
  \begin{subfigure}[b]{\textwidth}
    \centering
    \includegraphics[width=\textwidth]{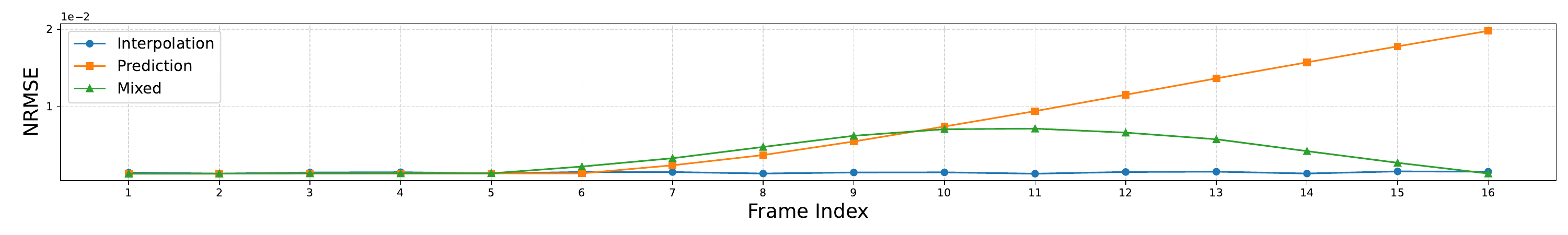}
    \label{fig:compare_error}
  \end{subfigure}
  \caption{Comparison of keyframe selection strategies: interpolation, prediction, and mixed. The top images display the reconstructed data for each strategy, while the bottom plot presents the per-frame NRMSE comparison across the three strategies. Interpolation-based keyframe selection outperforms the other two strategies.}

  \label{fig:keyframe_comparison}
\end{figure*}

\subsection{Compression Ratio}

The overall objective of scientific data compression is to achieve a high compression ratio while satisfying a user-defined error bound. Given a multi-dimensional scientific dataset \( \Omega \), a lossy compressor, and a corresponding decompressor, the compression process can be formulated as: $L = \text{Compressor}(\Omega)$, where \( L \) denotes the compressed representation. The reconstruction is obtained via: $\hat{\Omega} = \text{Decompressor}(L)$. To enforce the error bound, a post-processing step leveraging auxiliary data \( G \) is applied, resulting in the final reconstructed output \(\hat{\Omega}^G\). The effective compression ratio is then defined as:
\begin{equation}
    \text{Compression Ratio} = \frac{\text{Size}(\Omega)}{\text{Size}(L) + \text{Size}(G)}.
\end{equation}

\subsection{Evaluation Metrics and Datasets}

We employ the Normalized Root Mean Square Error (NRMSE) as a relative error criterion to evaluate the quality of reconstruction, taking into account that different datasets may span various data ranges. The NRMSE is defined in Equation \ref{eq:nrmse}:
\begin{equation}
\label{eq:nrmse}
\mathrm{NRMSE}\left(\Omega,\hat{\Omega}^{G}\right)=\frac{\sqrt{\left\|\Omega-\hat{\Omega}^{G}\right\|_{2}^{2}/N_d}}{\max\left(\Omega\right)-\min\left(\Omega\right)},
\end{equation}
where $N_d$ is the number of data points in the dataset, and $\Omega$ and $\hat{\Omega}^{G}$ represent the entire original dataset and the reconstructed datasets with error bound guaranteeing post-processing, respectively.



\paragraph*{\textbf{E3SM} Dataset}
The Energy Exascale Earth System Model (E3SM) \cite{e3sm} is a state-of-the-art computational framework designed to simulate Earth’s climate system with high resolution and accuracy. We evaluate our method using climate data produced by the high-resolution atmospheric configuration of E3SM. The simulation operates on a grid with 25~km spacing (equivalent to \(0.25^\circ\)), generating approximately 350{,}000 values per variable per hour. To convert the geospatial data into a format suitable for learning, we apply Cube-to-Sphere projections, mapping the Earth’s surface onto a planar grid. The resulting dataset has dimensions \(5 \times 8640 \times 240 \times 1440\), where 8640 corresponds to the number of hourly timesteps, 5 denotes the number of climate variables, and each frame has a spatial resolution of \(240 \times 1440\).

\paragraph*{\textbf{S3D} Dataset}

We now briefly introduce the S3D dataset \cite{s3d}, which represents the compression ignition of large hydrocarbon fuels under conditions relevant to homogeneous charge compression ignition (HCCI), as detailed in \cite{Yoo11}. The dataset comprises a two-dimensional space of size $512\times512$, collecting data over 200 time steps. A 58-species reduced chemical mechanism \cite{Yoo11} is used to predict the ignition of a fuel-lean $n$-heptane/air mixture. Thus, each tensor corresponds to 58 species, resulting in a spatio-temporal dataset with the shape $58 \times 200\times 512\times 512$.



\paragraph*{\textbf{JHTDB} Dataset}

We utilize a subset of the Johns Hopkins Turbulence Database (JHTDB) \cite{jhtdb}, a widely used resource offering direct numerical simulation (DNS) data for studying turbulent flows in scientific computing and turbulence research. Specifically, we use the isotropic turbulence dataset, which captures fully resolved turbulence with fine spatial and temporal granularity. The selected subset comprises a three-dimensional velocity field sampled on a uniform Cartesian grid with dimensions \(64 \times 256 \times 512 \times 512\), where 64 denotes the number of spatial regions, 256 indicates the number of time steps, and each spatial slice has a resolution of \(512 \times 512\).

\begin{table}[h!]
\centering
\begin{tabular}{c c c c c}
\hline
Application & Domain & Dimensions & Total Size \\ \hline
E3SM & Climate &  $5 \times 8640 \times 240 \times 1440$ & 59.7 GB\\ 
S3D & Combustion & $58 \times 200 \times 512 \times 512$ & 24.3 GB\\ 
JHTDB & Turbulence & $ 64 \times 256 \times 512 \times 512$ & 34.3 GB\\

\hline
\end{tabular}
\caption{Datasets Information}
\label{tab:dataset}
\vspace*{-0.5cm}
\end{table}

\subsection{Implementation Details}

Our experiments were conducted on the HiPerGator Supercomputer using single NVIDIA A100 GPUs for model training. The entire framework was implemented in PyTorch version 2.2.0. The model training process is divided into two stages: we first train the VAE with a hyperprior, followed by training the latent diffusion model. 

For VAE training, we randomly crop \(256 \times 256\) patches from the original data blocks. A batch size of 16 is used. For datasets with spatial dimensions smaller than 256 (e.g., E3SM), we apply reflection padding to reach the required size. The learning rate is initialized at \(1 \times 10^{-3}\) and decays by a factor of 0.5 every 100K iterations. The weight parameter \(\lambda\) is initialized to \(1 \times 10^{-5}\) and is doubled at the 250K iteration. Entire 500 K iteration is used for training. Scientific data typically exhibit a much larger range compared to natural images, often spanning from \( -10^{10} \) to \( 10^{10} \) in our applications. We normalize each frame independently to have zero mean and unit range.

To train the denoising latent diffusion model, we begin with $1,000$ denoising steps over 500K iterations, followed by fine-tuning with 32 steps for an additional 200K iterations. The training objective uses mean squared error (MSE) as the loss function. We set the batch size to 64 and the learning rate to \(1 \times 10^{-4}\). The temporal length \(N\) is set to 16, and each latent tensor has 64 channels.

\subsection{Keyframe Selection}



We investigate three keyframe selection strategies: prediction-based, interpolation-based, and mixed key framing. For the prediction-based strategy, the first 6 frames $\mathcal{C} = \{1, 2, 3, 4, 5,6\}$ are selected as keyframes for conditioning. The remaining frames are predicted by the model based on temporal correlations. In the interpolation-based strategy, 6 keyframes are uniformly sampled across the sequence $\mathcal{C}  = \{1, 4, 7, 10, 13, 16\}$, with the model conditioning on these frames to interpolate the missing ones. For the mixed strategy, the first 5 frames and the last frame are selected as keyframes $\mathcal{C} = \{1, 2, 3, 4,5, 16\}$, balancing early and final temporal context.

We visualize the original and reconstructed data in Figure \ref{fig:keyframe_comparison}, along with the frame-by-frame reconstruction error. Among these three keyframe selection strategies, the interpolation-based method shows the lowest reconstruction error compared to the other two methods. This is because the interpolation-based method provides a more gradual transition between frames, reducing the overall error. The closer the keyframes are to the conditioning frames, the lower the error in all strategies. This suggests that evenly distributed keyframes contribute to improved reconstruction accuracy.
\begin{figure*}[htbp]
    \centering
    \begin{subfigure}[b]{0.33\textwidth}
        \centering
        \includegraphics[width=\textwidth]{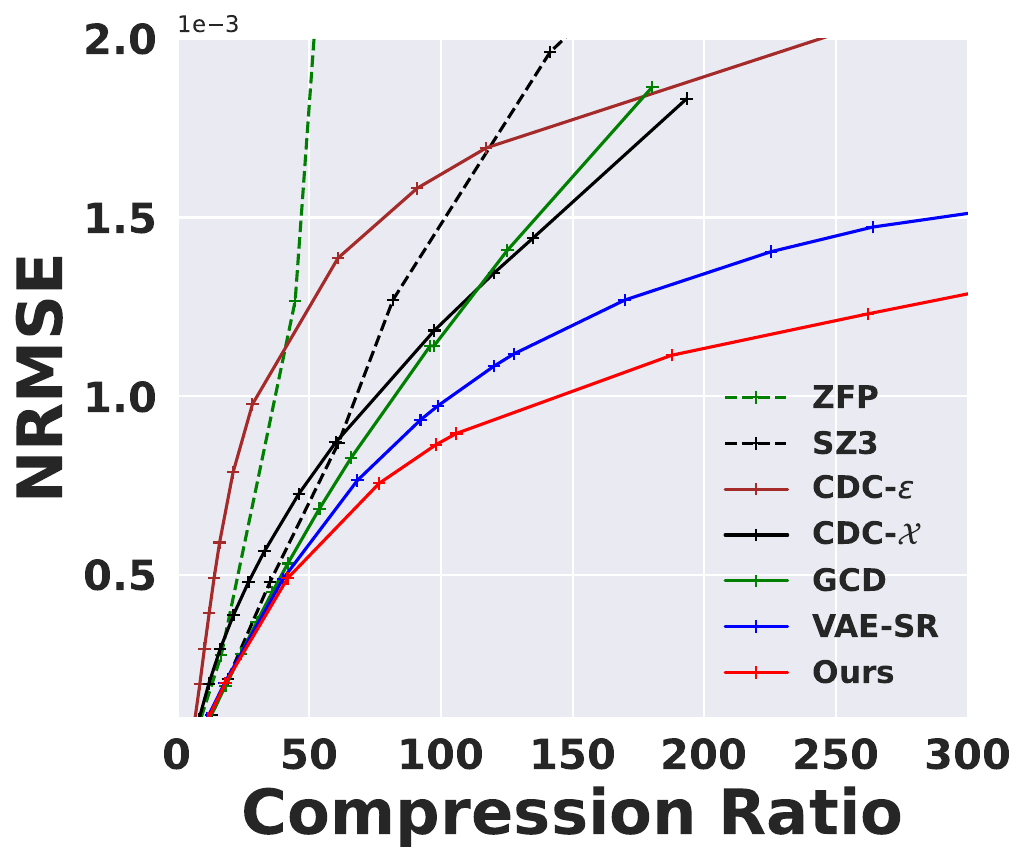}
        \caption{Evaluation on E3SM dataset.}
        \label{fig:e3sm_result_lcd}
    \end{subfigure}
    \begin{subfigure}[b]{0.33\textwidth}
        \centering
        \includegraphics[width=\textwidth]{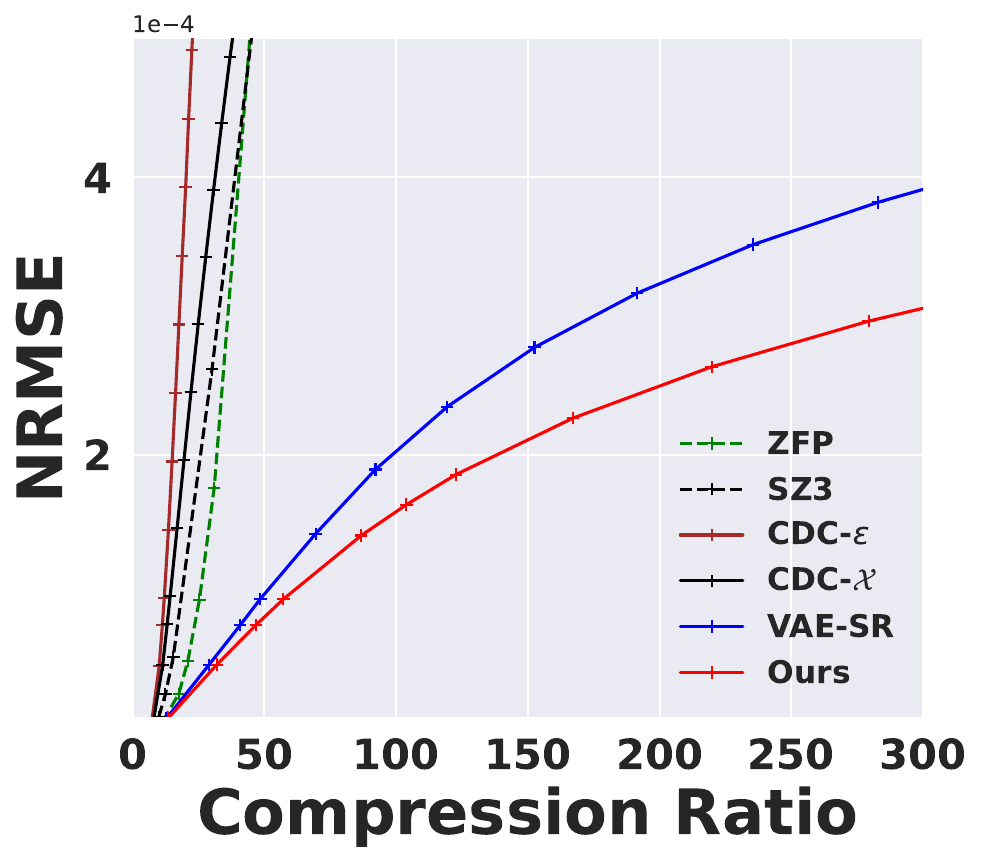}
        \caption{Evaluation on S3D dataset.}
        \label{fig:s3d_result_lcd}
    \end{subfigure}
    \begin{subfigure}[b]{0.33\textwidth}
        \centering
        \includegraphics[width=\textwidth]{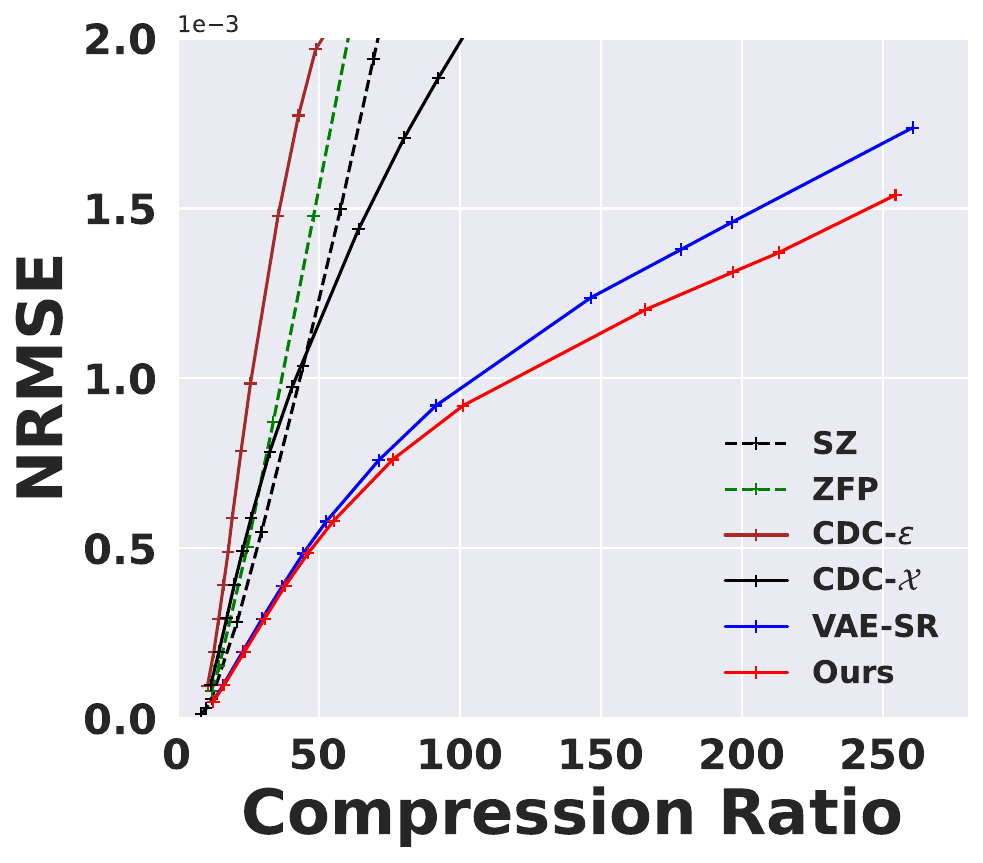}
        \caption{Evaluation on JHTDB dataset.}
        \label{fig:jhtdb_result_lcd}
    \end{subfigure}
    \caption{Comparison of results on the E3SM, S3D and JHTDB datasets.}
    \label{fig:main_compare}
\end{figure*}

\begin{figure}[b]
    \centering
    \includegraphics[width=0.5\textwidth]{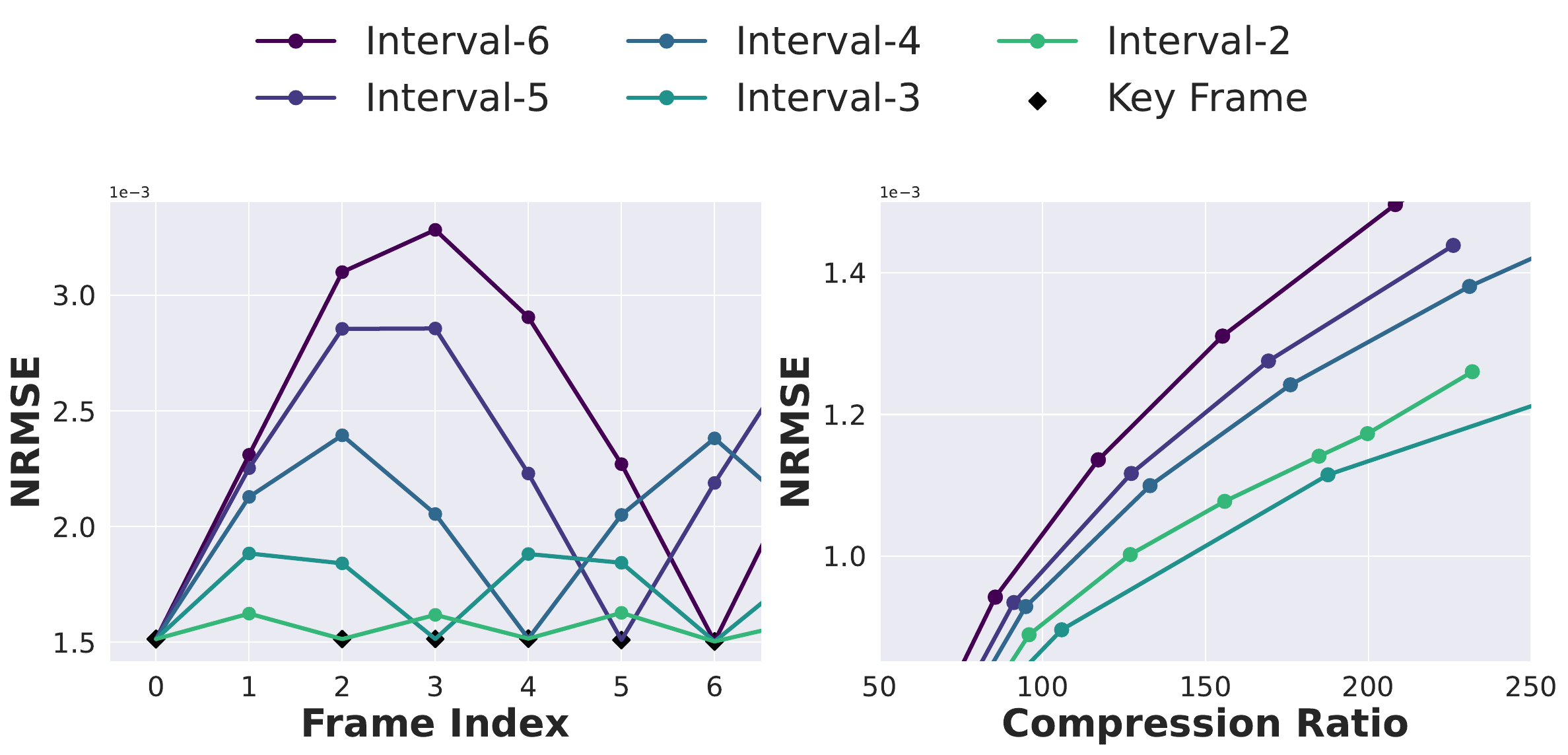}  
    \caption{Ablation Study for interpolation interval on E3SM dataset.}
    \label{fig:e3sm_interval}
\end{figure}

\subsection{Impact of Interpolation Interval}

For interpolation-based keyframe selection, increasing the keyframe interval generally leads to a higher compression ratio, but this may also elevate the reconstruction error. To investigate the trade-off between compression efficiency and reconstruction quality, we conduct experiments on the E3SM dataset, training identical models with keyframe intervals ranging from 2 to 6. After data reconstruction, we apply postprocessing with varying error bounds to obtain data points with different compression levels.

Figure~\ref{fig:e3sm_interval} (left) presents the per-frame NRMSE for each interval, while the right plot shows the NRMSE versus compression ratio curve. As anticipated, smaller intervals lead to lower reconstruction error due to more frequent keyframe conditioning. Although interval-2 results in the lowest reconstruction error, it incurs the highest storage cost. Interval-3 shows the best balance, offering an optimal trade-off between reconstruction accuracy and compression efficiency. Similar results are observed on the other two datasets, with the optimal keyframe interval being 3. While interval-3 performs best across our three applications, this interval may vary depending on temporal correlations and should be treated as a domain-specific hyperparameter.

\subsection{Impact of Denoising Steps}

Diffusion models typically suffer from slow generation speeds due to the need for thousands of denoising steps. While smaller denoising steps can speed up the process, they may make it difficult for the model to effectively learn useful features, often resulting in noisy outputs. In our application, training directly with smaller denoising steps leads to poor feature learning and noisy predictions. We found that training with larger denoising steps, followed by fine-tuning with smaller steps, achieves similar performance to training solely with large steps.

\begin{figure}
    \centering
    \includegraphics[width=0.3\textwidth]{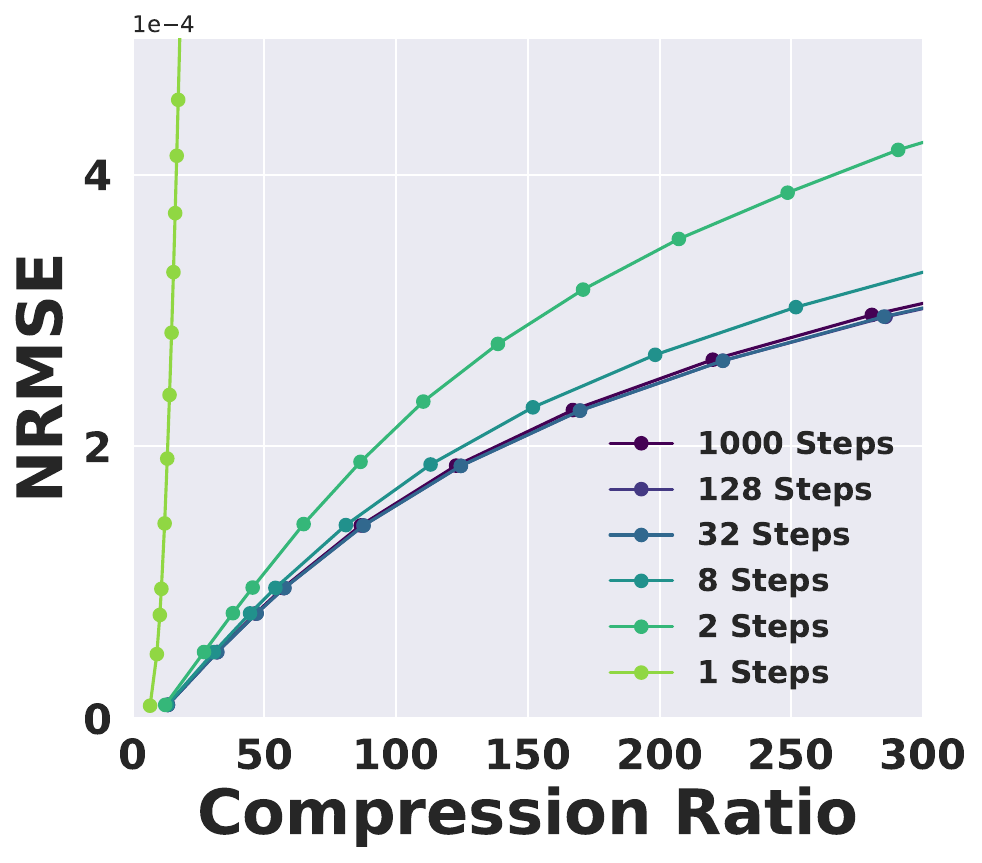}  
    \caption{Ablation Study for Denoising Step on S3D Dataset.}
    \label{fig:s3d_steps}
\end{figure}

As shown in Figure \ref{fig:s3d_steps}, we initially train each model with 1000 denoising steps and then directly fine-tune it with $\{128, 32, 8, 2, 1\}$ steps. While fewer denoising steps generally result in worse performance, we observe that using more than 32 steps achieves similar performance in terms of compression ratio and reconstruction accuracy compared to the original 1000 steps, while offering significantly faster performance.

\begin{figure*}[htbp]
    \centering
    \includegraphics[width=1\textwidth]{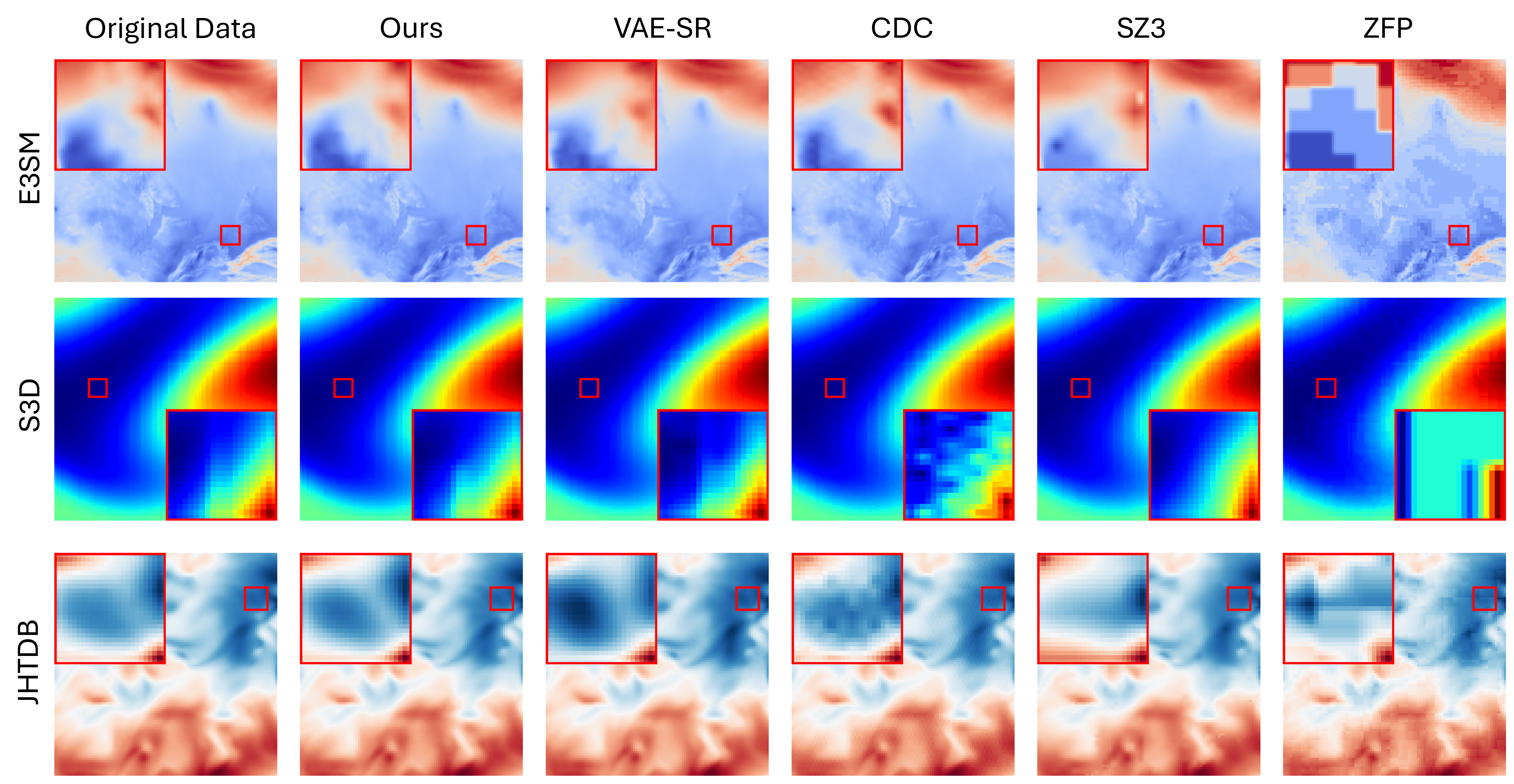}  
    \caption{Visualization of reconstructed data using our method, VAE-SR, CDC, SZ3, and ZFP compressors. All methods compress the data at a compression ratio of approximately 100. The two red rectangles indicate the same region—one shows the original location, while the other provides a zoomed-in view for detailed comparison.}
    \label{fig:vis_3domain}
\end{figure*}

\begin{table*}[htbp]
\centering
\caption{Inference Speed on GPU}
\begin{tabular}{ c c c c c c c c} 
\toprule
 Devices & Methods & Encoding Speed (MB/s) & Decoding Speed (MB/s)
\\
\midrule
\multirow{6}{5em}{A100 80GB}
&CDC-$\mathcal{X}$& 850& 4 \\
&CDC-$\epsilon$ & 907& 1.4 \\
&GCD &  605 & 0.3 \\
&Ours-128 Steps & 1942& 16 \\
&Ours-32 Steps &  1942 & 61 \\
&Ours-8 Steps & 1942 & 203 \\
\bottomrule

\multirow{6}{5em}{RTX 2080 24GB}
&CDC-$\mathcal{X}$& 282& 1.1 \\
&CDC-$\epsilon$ & 191& 0.4 \\
&GCD &  374 & 0.2 \\
&Ours-128 Steps & 825 & 6\\
&Ours-32 Steps & 825 & 25\\
&Ours-8 Steps & 825 & 93\\

\bottomrule

\end{tabular}

\label{tab:speed}
\end{table*}

\subsection{Comparison with Other Methods}

We evaluate our method on three benchmark datasets and compare it against SZ3 \cite{sz3}, ZFP \cite{zfp}, CDC \cite{cdc}, GCD \cite{gcd}, and VAE-SR \cite{xiaofm}. SZ3 and ZFP are two rule-based state-of-the-art scientific data compressors, where SZ3 utilizes a prediction-based approach and ZFP relies on transform-based compression. CDC represents a leading method in image compression based on conditional diffusion models, where CDC-$\mathcal{X}$ predicts the original signal directly, and CDC-$\epsilon$ predicts the noise added during the diffusion process. To apply CDC to 3D data, we treat three consecutive frames as a three-channel input. GCD extends CDC from 2D to 3D, enabling it to handle spatiotemporal datasets effectively. VAE-SR employs a variational autoencoder to encode the data and integrates a super-resolution module to enhance the fidelity of the reconstructed output.

As shown in Figure \ref{fig:main_compare}, our method achieves substantial compression improvements across all three datasets compared to the listed baselines. In the figure, rule-based methods (ZFP and SZ3) are represented with dotted lines, while learning-based methods are shown with solid lines. We observe that learning-based methods generally achieve significantly higher compression ratios at the same reconstruction error. This is because rule-based compressors, which are designed for general purposes, often struggle to capture domain-specific features and correlations, limiting their compression capabilities

Compared to the best rule-based method, SZ3, our approach achieves up to 4$\times$ higher compression on the E3SM dataset, 10$\times$ higher compression on S3D, and 5$\times$ higher compression on JHTDB. When compared to the strongest learning-based baseline, VAE-SR, our method achieves up to a 63\% improvement in compression ratio on E3SM, 62\% higher compression on S3D, and 20\% higher compression on JHTDB. One key advantage of our approach is that we only need to store the latent representation of keyframes, while CDC, GCD, and VAE-SR must retain latent representations for every frame or every data block. This significantly reduces storage overhead. Our results highlight the effectiveness of generative models in scientific data compression by leveraging learned priors to reconstruct non-keyframes with high fidelity.

\subsection{Inference Speed}

To demonstrate the inference efficiency of our method among generative approaches, we compare it against two representative generation-based compression methods: CDC and GCD. Evaluations are conducted on both an NVIDIA A100 80GB GPU and an NVIDIA RTX 2090 24GB GPU to assess performance across different hardware configurations. We report encoding and decoding times separately to provide a more detailed breakdown of computational cost.

All three methods share a common design pattern: the encoder is relatively lightweight, while the decoder is more computationally intensive due to the generative process. Specifically, the diffusion process occurs during decoding. However, a key difference lies in the domain where diffusion is applied—CDC and GCD perform diffusion directly in the original data space, which is high-dimensional and computationally expensive. In contrast, our method performs diffusion in the latent space, which is significantly more compact and efficient.

As shown in Table~\ref{tab:speed}, this design choice significantly reduces the decoding time of our method while maintaining high reconstruction quality. With 32 denoising steps, our method outperforms CDC by over 2× in encoding speed and 15× in decoding speed on NVIDIA A100 GPU. It also surpasses GCD, achieving more than 3× faster encoding and 200× faster decoding. Beyond speed, our method delivers higher compression, making it particularly well-suited for large-scale or real-time scientific data applications.

The comparison in this section focuses exclusively on model inference time. The overhead from enforcing the error bound guarantee is excluded, as all three methods could rely on the same postprocessing procedure. Currently, this step is implemented on the CPU, making it slower than GPU-based model inference. We are actively working on accelerating this component through GPU optimization.

\section{Conclusion}
In this work, we propose a generative compression framework that combines a VAE-based transform with a latent diffusion model to efficiently reduce the storage cost of scientific spatiotemporal data. We compress only selected keyframes using a VAE with a hyperprior and reuse their latent representations as conditioning inputs to guide a latent diffusion model that reconstructs intermediate frames. To improve efficiency, the diffusion model is first trained with a large number of denoising steps and then fine-tuned with fewer steps, significantly accelerating inference. Our approach achieves up to 10× higher compression compared to rule-based methods like SZ3 and up to 63\% improvement over leading learning-based compressors. We apply a post-processing module based on PCA to ensure strict error-bound guarantees, making the approach suitable for scientific workflows where the accuracy of primary data is critical. Looking forward, we aim to further improve the practicality of our approach by accelerating the post-processing module using GPU implementations, which would enable a fully GPU-based end-to-end compression and decompression pipeline.


\bibliographystyle{ACM-Reference-Format}
\bibliography{sample-base}

\appendix

\end{document}